# MicroAnalyzer: A Python Tool for Automated Bacterial Analysis with Fluorescence Microscopy

Jonathan Reiner[1,¶], Guy Azran[1,¶], Gal Hyams[1]

[1]The Department of Computer Science and Engineering, The Hebrew University of Jerusalem, Israel

[¶]These authors contributed equally to this work.

*Abstract* — **Fluorescence microscopy is a widely used method among cell biologists for studying the localization and co-localization of fluorescent protein. For microbial cell biologists, these studies often include tedious and time-consuming manual segmentation of bacteria and of the fluorescence clusters or working with multiple programs. Here, we present MicroAnalyzer - a tool that automates these tasks by providing an end-to-end platform for microscope image analysis. While such tools do exist, they are costly, black-boxed programs. Microanalyzer offers an open-source alternative to these tools, allowing flexibility and expandability by advanced users.**

**MicroAnalyzer provides accurate cell and fluorescence cluster segmentation based on state-of-the-art deep-learning segmentation models, combined with ad-hoc post-processing and Colicoords - an open source cell image analysis tool for calculating general cell and fluorescence measurements. Using these methods, it performs better than generic approaches since the dynamic nature of neural networks allows for a quick adaptation to experiment restrictions and assumptions. Other existing tools do not consider experiment assumptions, nor do they provide fluorescence cluster detection without the need for any specialized equipment.**

**The key goal of MicroAnalyzer is to automate the entire process of cell and fluorescence image analysis "from microscope to database", meaning it does not require any further input from the researcher except for the initial deep-learning model training. In this fashion, it allows the researchers to concentrate on the bigger picture instead of granular, eye-straining labor.**

## I. INTRODUCTION

### 1.1. Background

Recent advancements in fluorescence microscopy allow researchers to detect proteins within single-celled microorganisms and to determine their specific subcellular localization (different patterns of localizations can be seen in [1] Figure 1), providing new insights into numerous molecular processes [2]–[6]. The endcaps of rod-shaped bacterial cells, termed poles, emerge as hubs for protein clusters [7], [8].

Protein localization studies can be performed using microscopic image analysis. In order to produce the images, the researchers must:

- Grow bacteria with fluorescence proteins bound to the proteins being studied.
- Carefully and evenly place the bacteria on a petri dish, perhaps using an adhesive material in order to keep the cells in place.
- Taking multiple images of different regions on the petri dish.

Advancements in fluorescence microscopy automation over the past decade have given researchers the ability to produce thousands of images overnight [9]–[11], creating a demand for fast and reliable microscopic image analysis on the bacterial images and their fluorescence channels. Current methods for performing these tasks involve using specialized programs that allow manual segmentation of the cells and fluorescence clusters for every image, then performing automated analysis using the segmentations as a baseline for their location and general shape [12], [13]. While part of the segmentation process may be automated as well, the used algorithms are not reliable enough to allow their outputs to go unchecked. This results in a time-consuming process that requires much human interaction and decision making.

A common solutions for studying fluorescence localization include advanced super-resolution microscopy (SR) [14] techniques based on fluorescence photoactivated localization microscopy (FPALM) [15], such as Stochastic optical reconstruction microscopy (STORM) [16], that can obtain spatial data on single fluorescent molecules. While this technique simplifies the issue of cluster segmentation immensely, SR capable microscopes can be very expensive and are not available in every lab.

Open-source developers and some private companies have been attempting to fully automate the process of cell and cell nuclei segmentation using deep-learning techniques [17], [18]. However, these algorithms attempt to generalize this task to many different bacterial species and try to find as many cells as possible, meaning they do not take into consideration experiment constraints, such as the researchers' preferred cell size and or desired spacing. Furthermore, none of them [17], [18] handle segmentation of fluorescence clusters which is required for calculating localization metrics of the observed material.

After the information has been extracted from the images (cells and clusters segmentation), the researchers must perform calculations on that data. This information must be accurate in order to ensure the reliability of collected statistics. While there are existing tools that perform this analysis, the open-source (free) options are standalone, i.e. do not contain the segmentation feature. This requires the researchers to move data from one tool to another, and they must do so carefully to avoid data corruption.

MicroAnalyzer attempts to be the solution to the following problem: **how can the process of**

**analyzing [1] rod-shaped cells and polar fluorescence clusters in a raw image from a microscope be automated?** This automation task can be divided into three subtasks:

i. *Cell segmentation* – finding a *good enough*[2] algorithm to get a rough estimation of the location of the cells under the restrictions of the experiment.

ii. *Fluorescence clusters segmentation* – finding *a good enough* algorithm to find the location of clusters within cells.

iii. *Cell and fluorescence analysis* – finding accurate locations and measurements of cells and fluorescence clusters using the segmentation results acquired in the previous tasks.

1.2. **Experiment Assumptions.**

The article follows the requirements for a set of experiments conducted by Orna Amster-Choder's lab and thus takes their main assumptions:

a) Cells that are *too* close together are invalid and should not be analyzed.

b) Cells that are *out of focus* are invalid and should not be analyzed.

c) Minimize false positive cell detections (a false positive cell detection is worse than a false negative).

d) Fluorescence clusters that do not intersect with the boundaries of a cell should not be analyzed and are to be viewed as noise.

MicroAnalyzer, is a tool that accepts the lab's microscope's raw image files and outputs a full analysis database including fluorescence channel data, under the above assumptions with useful visualizations.

In order to evaluate MicroAnalyzer's results, this paper introduces a new criterion for segmentation model validity. This criterion takes into consideration the experiment's assumptions and the possibility of multiple ground truths as a result of disagreement between researchers.

1.3. **Prior Solutions**:

There are several available tools for solving cell segmentation, fluorescence cluster segmentation and cell and fluorescence analysis. Some of them provide an end-to-end platform for all sub-tasks, and some solve only a single sub-task.

Existing tools that offer the entire pipeline, from the raw microscope image to the final output database, such as ImageJ [12], require the user to perform the segmentation for cells and fluorescence clusters manually. This is a time-consuming and error prone task. Private companies, Nikon for example, offer proprietary, paid programs (e.g. NIS-Elements [13]) that have similar features to ImageJ, but offer deep learning algorithms that can perform the segmentation task without human interaction (after training) as a paid plugin. These are closed source tools and are therefore not expandable.

---

[1] "Analysis" refers to performing calculations for the output database (see appendix F)

[2] A "good enough" algorithm is one that provides "valid predictions" as defined in section 3.1.

Other open source alternatives offer solutions to each of the sub-tasks separately, and do not perform the entire task (from microscope to database) in an automated fashion [17]–[19].

Cell and cluster segmentation can be achieved by using different neural network architectures for segmentation and object detection [20]–[26]. They offer a general solution for these segmentations which do not require any manual configuration (after training) and have been found reliable for similar tasks.

## II. MICROANALYZER METHODS

### 2.1. Components.

MicroAnalyzer consolidates open source solutions for each subtask into a single tool that performs the entire pipeline of operations: cell segmentation, fluorescence cluster segmentation and data analysis (*Figures 1* describe the flow of operations performed in the program). The chosen methods for each task are:

i) Cluster segmentation – Feature Pyramid Network (FPN) [27], a segmentation neural network..

ii) Cell segmentation - Mask-RCNN [28], an object detection neural-network.

iii) Cell and fluorescence analysis – The cell analyzing component of MicroAnalyzer (CellAnalyzer) is a modified version of Colicoords (see 1.3), that supports cluster segmentation data and calculations.

FPN is an object segmentation convolutional neural network that uses a special architecture in order to observe the data at different resolutions, i.e. different detail levels, similar to the idea behind the U-Net segmentation network [23]. This architecture keeps the image at multiple resolution levels and maintains strong semantic features throughout these levels, giving it an edge in segmenting smaller objects over its predecessors

Mask-RCNN is a state-of-the-art object detection neural network that performs instance segmentation at the object-level. This is done by initially finding regions where the location of an object is suspected, then classifying the object in that region, and finally finding a pixel-wise

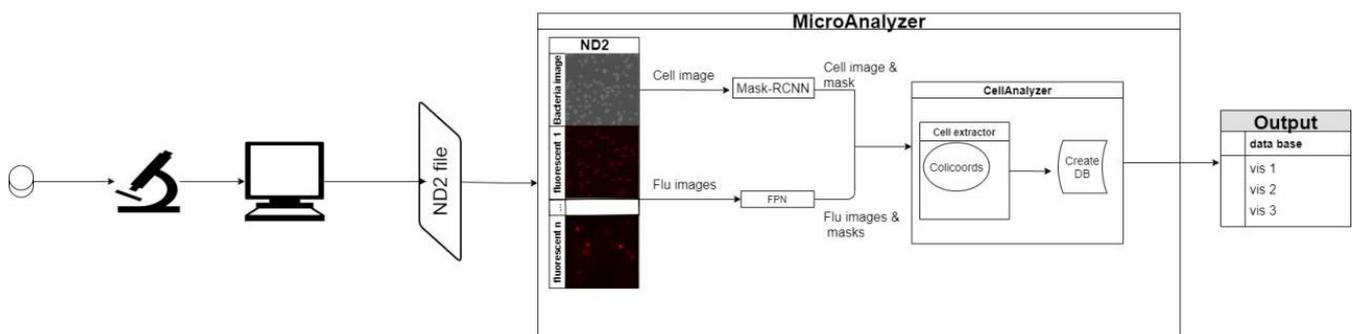

*Figure 1*: *The chart describes the full flow of MicroAnalyzer form microscope to database. The microscope takes images and outputs a computer readable ND2 file which is input for MicroAnalyzer. It performs segmentation on cells and fluorescence clusters using Mask-RCNN and FPN networks respectively, performs analysis to create using the CellAnalyzer module and finally output the database.*

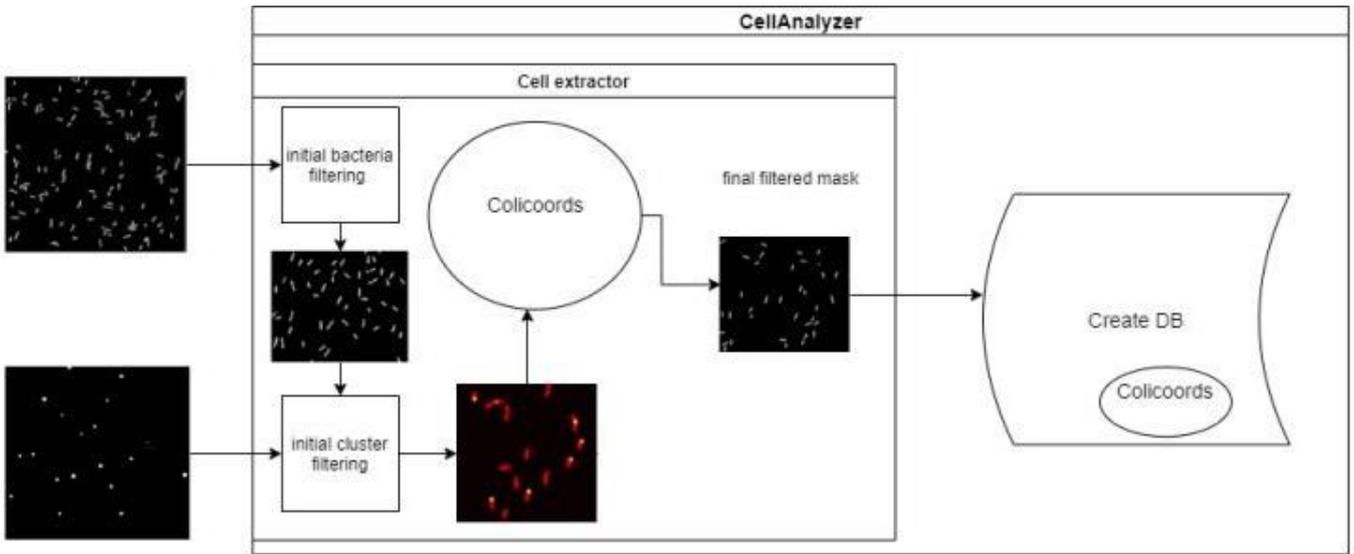

*Figure 2*: *CellAnalyzer module flow chart for cell and fluorescence cluster analysis. It performs filtering on both clusters and bacteria based on minimum cell dimensions and proximity and cluster intersection with cells. Colicoords is used to extract the individual cells from the image fits a coordinate system for them, creating the final and accurate binary mask. The extracted cell information is used to calculate and populate the database fields.*

segmentation of each object. More accurately, this model attempts to minimize three losses simultaneously:

- $L_{box}$ – bounding-box regression (defined in [29] appendix C).
- $L_{cls}$ – classification loss (average categorical cross-entropy loss).
- $L_{mask}$ – mask loss (average binary cross-entropy loss).

This is in fact an evolution of "Faster RCNN" [25] for finding the regions of interest using an FPN backbone for pixel-wise segmentation.

Colicoords is an open source bacterial image analyzer that structures microscopy data at the single-cell level and implements a coordinate system describing each cell [19]. Given a microscope image of rod-shaped cells and their fluorescence channels and a segmentation of some or all of the cells in the image, Colicoords fits the aforementioned coordinate system for each cell according to the original microscope image using the provided segmentation as a baseline. This coordinate system can also be used to map the exact location of the cell in any and all fluorescence channels, allowing for calculations on those channels within the cell boundary.

Using the above tools, along deterministic calculations on the fluorescence clusters (not included in Colicoords), allows the user to perform high accuracy end-to-end cell and fluorescence cluster analysis in a fully automated environment while being free and open source. Moreover, the code is highly tunable and adaptable to other experiments and conditions.

2.2. **Pipeline**

*Figure 1* shows the entire flow of operations for a study using MicroAnalyzer. After the microscope has finished a photo session, an output ND2 file is created. This is a Nikon proprietary binary file containing all the camera's channels, including the grayscale bacteria image and the fluorescence intensity channels. given such a file, MicroAnalyzer initially separates the bacteria image channel from the fluorescence channels.

The images are pre-processed (see appendix C) and fed into their respective models for segmentation / detection (Mask-RCNN for cells, FPN for clusters). The models' outputs are binary masks indicating the location of cells / clusters in the image.

The output masks are sent to the CellAnalyzer module (see *Figure 2*) which filters invalid cells and fluorescence clusters using deterministic algorithms based on minimal object size and proximity. All images and their corresponding masks are analyzed by Colicoords, which accurately extracts cell information with the underlying fluorescence data. Using this output, CellAnalyzer is able to map fluorescence clusters to their enclosing cells and then calculate desired database fields and construct the output database and visualizations.

## III. EVALUATION METHODS

### 3.1. Segmentation

The responsibility of the segmentation phase is to find the general location and shape of the cells and fluorescence clusters (since the accurate location and shape are found in the analysis phase). This is represented by a binary image (of ones and zeros) where the ones represent pixels where a cell can be found in the input image and zeros represent the background.

In experiments with non-deterministic assumptions, one must consider the possibility of having more than one ground truth segmentation (see *Figure 3*. For example, assumptions (a) and (b) are subject to researcher variability since the definitions "too close together" and "out of focus" are open to interpretation. One researcher may find a cell acceptably sharp in an image while another may decide it to be blurry and omit it from the final segmentation.

There are many possible ways to evaluate the quality of a segmentation, such as the classic metrics, including accuracy, precision, recall, etc., metrics that combine the classic metrics, e.g. f-score, and specialized metrics, e.g. Jaccard loss / IoU score. All of the above metrics are measured on a pixel level, meaning that they penalize a segmented object if it is slightly smaller or larger than the ground truth segmentation. Since the task only requires finding the general location and shape of the objects[3], these metrics have less meaning in this scenario.

Alternatively, one can view this problem as an Object Detection task. In that case, common evaluation metrics are average precision (AP), average recall (AR), average $f_\beta$-score (AF), etc., based on intersection over union score (IoU) thresholding. These metrics fall short as well – most of them cannot not take into account the experiment assumptions, e.g. penalize more on false positives, and none of them consider the possibility that researchers may provide very different segmentations for the same image (see *Figure 3*). The evaluation metric defined in the next paragraphs takes a similar approach to AF in the sense that it uses an IoU threshold to identify false positive and negative detections, but takes into account both this experiment assumptions and multiple ground truth possibilities.

---

[3] A crude binary mask of cell / fluorescence clusters that tells Colicoords where to search for cells.

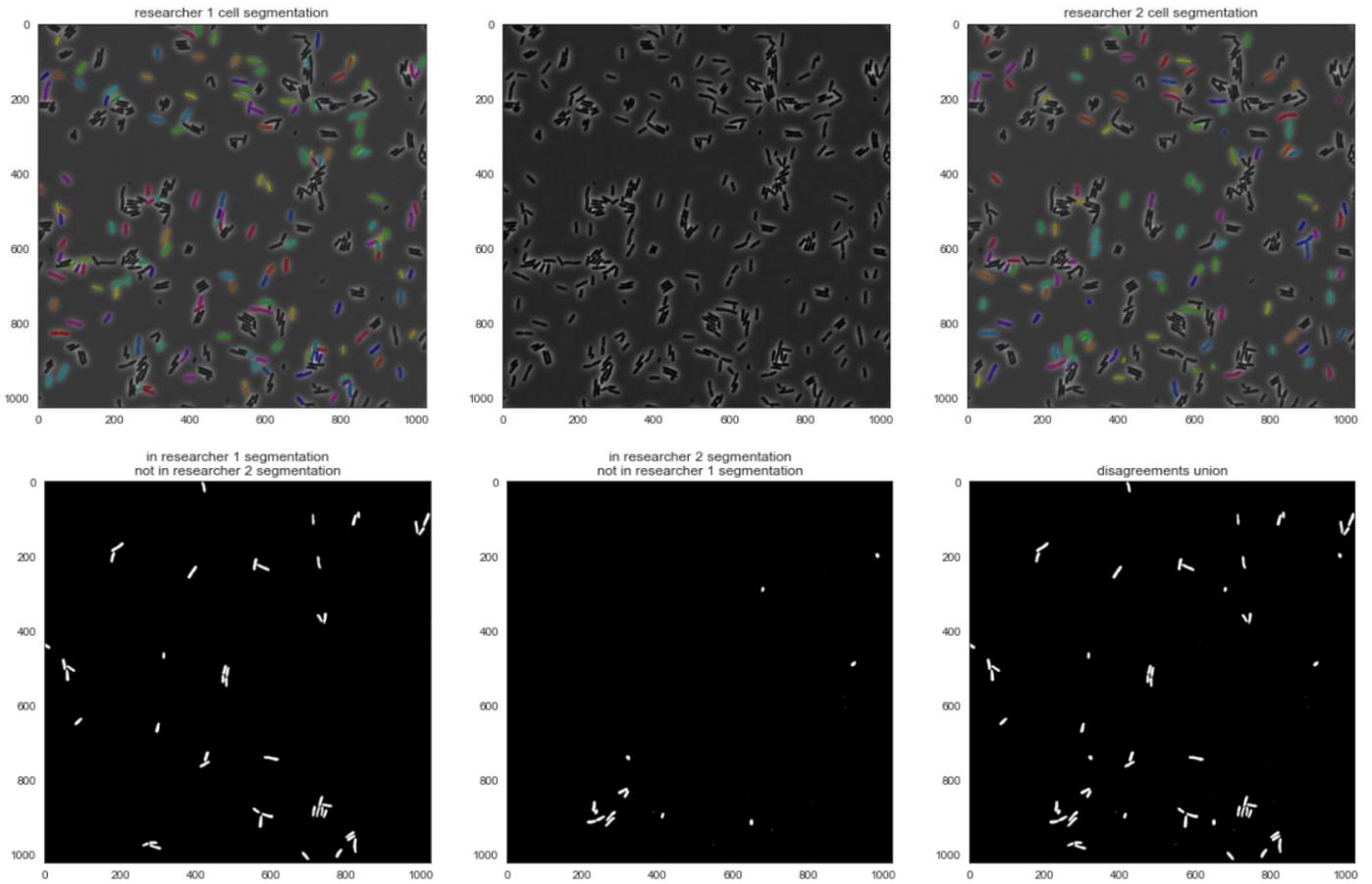

*Figure 3: An example of two different ground truths for the same image. The top row presents the cell segmentation of each researcher (cells are randomly colored to divide nearby cells) where the left and right images show segmentations of different researchers. The bottom row shows the differences between the segmentations.*

### 3.1.1. Intersecting Connected Components

Let $s_1, s_2 \in M_{n \times m}(\{0,1\})$ be two binary segmentations of the same image. A *connected component* in binary segmentation $s$ is a set $c$ of pixel coordinates, i.e. $c \subseteq n \times m$, such that

- for all $(i,j) \in c$, $s_i[i,j] = 1$
- all neighboring coordinates $(k,l) \in n \times m$ $(= (i \pm 1, j), (i, j \pm 1))$ have pixel value 1 if and only if $(k,l)$ is part of the connected component, i.e.:

$$s_i[k,l] = 1 \Longleftrightarrow (k,l) \in c$$

Define $C_s$ to be the set of connected components in $s$. It is possible, and even intuitive, to only consider connected components since all cells and fluorescence clusters appear connected in the images.

Given connected components $c_1, c_2$, the IoU score is defined to be:

$$IoU(c_1, c_2) = \frac{|c_1 \cap c_2|}{|c_1 \cup c_2|}$$

Say that they *intersect* if $IoU(c_1, c_2) \geq T$ where $T$ is a predefined threshold. For this experiment, choose $T_{Cell} = 0.8^4$, and $T_{fluo} = 0.6$.

Let $T > 0.5$ be the chosen threshold and $a, b \in C_{s_1}$, $c \in C_{s_2}$ and assume that $IoU(a, c) > T$ and $IoU(b, c) > T$. Then by the pigeonhole principal, $a \cap b \neq \emptyset$ and since these are connected components in the same segmentation then $a = b$. This defines an equivalence relation given two binary segmentations $a, b \in C_{s_1} \cup C_{s_2}$:

$$a \overset{\{s_1, s_2\}}{\sim} b \Leftrightarrow IoU(a, b) > T$$

Finally, define

$$C_{[s_1, s_2]} = \left\{ [c]_{\overset{\{s_1, s_2\}}{\sim}} \middle| c \in s_1 \right\}$$

In other words, $c_1 \in C_{[s_1, s_2]}$ and $c_2 \in C_{[s_2, s_1]}$ are equal if and only if $IoU(c_1, c_2) > T$.

Let $FP$ be a function such that $FP(s_1, s_2) = |C_{[s_1, s_2]} \setminus C_{[s_2, s_1]}|$ is the number of connected components in $s_1$ that do not intersect with any connected components in $s_2$, i.e. extra objects in $s_1$ that do not appear in $s_2$. These are the false positive predictions in $s_1$ given that $s_2$ is the ground truth.

Let $FN$ be a function such that $FN(s_1, s_2) = |C_{[s_2, s_1]} \setminus C_{[s_1, s_2]}|$ is the number of connected components in $s_2$ that do not intersect with any connected components in $s_1$, i.e. missing objects in $s_1$ that appear in $s_2$. These are the false negative predictions in $s_1$ given that $s_2$ is the ground truth.

### 3.1.2. Experimental $l_{ex}$-Error

The vagueness of this experiment's assumptions, e.g. assumption (a) uses the term *too close*, can lead severe differences between two researchers' segmentations on the same image as one researcher might be more conservative while the other might be fairly permissive. To account for this, there must be more than one researcher segmentation. Let $G_1$ and $G_2$ be two possible ground truth segmentations for the same image.

Given one of the ground truth binary segmentations $G_i$, a prediction binary segmentation $pd$, and $\beta \in [0,1]$, define the experimental $l_{ex}$-error of $pd$ according to $G_i$ to be:

$$l_{ex}(pd, G_i) = \frac{\beta \cdot FP(pd, G_i) + (1 - \beta)FN(pd, G_i)}{|C_{[G_1, G_2]} \cup C_{[G_2, G_1]}|}$$

This score attempts to avoid the classic metric issue of pixel-wise loss/scoring while also taking assumption (c) into consideration by penalizing more on extra objects (false positives) than missing objects using large $\beta$ value.

One might want to consider using AF instead of $l_{ex}$-error. However, using AF or other similar metrics, e.g. AP and AR, leads to unintuitive results in terms of the experiment assumptions. For example, according to assumption (c) it is better to detect no cells in the image than to detect *many* false positive cells, but the AP, AR and AF for a blank mask will always be 0 (since it does not find any true positive detections) and detecting all the cells in an image (such that many of them are not valid for this experiment) will receive a positive score, indicating that it is a better prediction than the blank mask (see *Table 1*).

---

[4] Lower thresholds accepted cell masks that did not provide enough context to Colicoords, causing runtime errors and in worse cases invalid output data.

### 3.1.3. Experimental Distance

The experimental distance between them is defined as:

$$d_{ex}(G_1, G_2) = \frac{l_{ex}(G_1, G_2) + l_{ex}(G_2, G_1)}{2}$$

i.e. the average experimental $l_{ex}$-error of considering each segmentation as the ground truth. Simplifying reveals that:

$$\frac{l_{ex}(G_1, G_2) + l_{ex}(G_2, G_1)}{2}$$
$$= \frac{1}{2}\big(\beta FP(G_1, G_2) + (1-\beta)FN(G_1, G_2)$$
$$+ \beta FP(G_2, G_1)$$
$$+ (1-\beta)FN(G_2, G_1)\big)$$
$$= \frac{1}{2}\big(\beta |C_{[G_1,G_2]} \setminus C_{[G_2,G_1]}|$$
$$+ (1-\beta)|C_{[G_2,G_1]} \setminus C_{[G_1,G_2]}|$$
$$+ \beta |C_{[G_2,G_1]} \setminus C_{[G_1,G_2]}|$$
$$+ (1-\beta)|C_{[G_1,G_2]} \setminus C_{[G_2,G_1]}|\big)$$
$$= \frac{1}{2}\big(|C_{[G_1,G_2]} \setminus C_{[G_2,G_1]}| + |C_{[G_2,G_1]} \setminus C_{[G_1,G_2]}|\big)$$
$$= \frac{1}{2}|C_{[G_1,G_2]} \Delta C_{[G_2,G_1]}|$$

Basically, the experimental distance is proportional to the disagreed objects in the segmentations, i.e. the number of objects that appear in exactly one of the segmentations. This gives a representation of the similarity of two segmentations in the task's context.

### 3.1.4 Valid Predictions

Given a prediction segmentation $pd$ and two ground truth segmentations $G_1, G_2$, $pd$ is called a *valid* prediction if:

$$\frac{l_{ex}(pd, G_1) + l_{ex}(pd, G_2)}{2} \leq d_{ex}(G_1, G_2)$$

Ultimately, a valid prediction is one that is less "far away" from the ground truth segmentations than they are from each other, meaning the prediction might as well be the segmentation of another researcher, i.e. another ground truth segmentation.

For this experiment, in order to satisfy assumptions (a-c), define $\beta_{cell} = 0.7$ in order to penalize false positive predictions more and attempt to minimize the segmentation of invalid cells that negate these assumptions. Also define $\beta_{fluo} = 0.15$ which will encourage to choose a model that finds as many fluorescence clusters as possible while missing as little as possible ground truth clusters. Using assumption (d) and the already generated cell segmentation, segmented clusters that are not within a cell boundary can be later removed. That is why in the fluorescence cluster segmentation task the goal should be to find many valid clusters, even if they are not within cells.

### 3.2. **Analysis**.

The analysis phase calculates the measurements of cells and fluorescence intensity using

Colicoords. This includes fields[5] C – J in the database. Fields K – T were calculated in the methods defined by the researchers and are described in appendix F (API).

## IV. RESULTS

MicroAnalyzer expects to receive images of cells with their fluorescence channels. In this experiment, the data was received as '.nd2' files, a Nikon proprietary binary file type (handled using an open-source solution nd2reader [30]).

The given data[6] for evaluation was 45 cell images and 31 fluorescence images of size $1022 \times 1024$. With each cell or fluorescence image was provided a ground truth binary mask from one of the experiment's researchers generated manually using NIS-Elements. According to these ground truth segmentations, the cell images have an average of 38.8 segmented cells per image and the fluorescence images have an average of 57.5 segmented clusters per image. The dataset of cell images was divided into a training set consisting of 40 images and a test set consisting of 5 images. The dataset of fluorescence images was split into a training set consisting of 27 images and a test set consisting of 4 images.

Additionally, two other images were segmented by two different researchers independently from one another. These two images will be called the validation set[7] and denoted $\{G_{i,j}\}_{(i,j) \in 2 \times 2}$ where $G_{i,j}$ is the ground truth segmentation of image $i$ by researcher $j$. All models are evaluated on the validation set.

Images were all taken with the same microscope. All cells that appear in the images are rod-shaped. The fluorescence channels contained several types of proteins and RNA.

### 4.1. Cell Segmentation.

In this section, two main approaches were taken in order to perform the segmentation. The first is a deterministic approach which uses thresholding-based techniques to find the ideal algorithm for valid cell segmentation. The second is the use of known segmentation and detection neural networks in order to attempt to mimic a researcher.

#### 4.1.1. Thresholding-based algorithm.

Thresholding was performed using several existing methods, including minimum [31] and Yen [32] thresholding. Using such algorithms, it is possible to find cell-shaped objects in the input images. Some outputs do require further processing, but all methods basically find the same set of cells in the images. Nonetheless, these methods with their default settings find all (or most) of the cell shapes in the image (see *Figure 4*) and do not take into account any of the experiment assumptions. This includes fuzzy, out of focus cells and extremely crowded cells. Correspondingly, the experimental $l_{ex}$-error is very high. Furthermore, inputting the segmentation (that contains all cells in the image) into MicroAnalyzer's CellAnalyzer

---

[5] See appendix F (API)
[6] Can be downloaded via the link in appendix A (Dataset)

[7] Validation images and the evaluation process can be found in appendix A under the link "Validation"

module does not filter the prediction enough to make it valid i.e.

$$\frac{l_{ex}(pd, G_{i,1}) + l_{ex}(pd, G_{i,2})}{2} > d_{ex}(G_{i1}, G_{i2})$$

*Figure 5* shows the resulting cell mask of using minimum thresholding and CellAnalyzer post-processing.

It is possible to achieve better results using custom configurations, but the acceptable configurations change from image to image. This is the opposite of automation, and thus this method is not used in MicroAnalyzer, which aims to request as little information as possible from the user.

Note: Maybe correct thresholding configurations can be learned using known machine learning methods.

4.1.2. **Deep-learning algorithm**

Three known neural network models were tested for this task: U-Net, Mask RCNN, and Cellpose. The first two were trained in a similar manner, and the Cellpose model was trained in the restricted settings of its package.

U-Net [23] is a convolutional neural network originally created for biomedical segmentation. It outperformed its predecessors greatly by introducing a new approach different from the then popular "sliding window" method. The idea behind this network is to convolve over and down-sample the image several times, using more filters as the image gets smaller, training it to slowly reduce the image to context information. This information is then up-sampled and combined with the different image resolutions and then reduced via convolutions into the final segmentation. For its purpose and ground-breaking performance of its time, it is used as a baseline model.

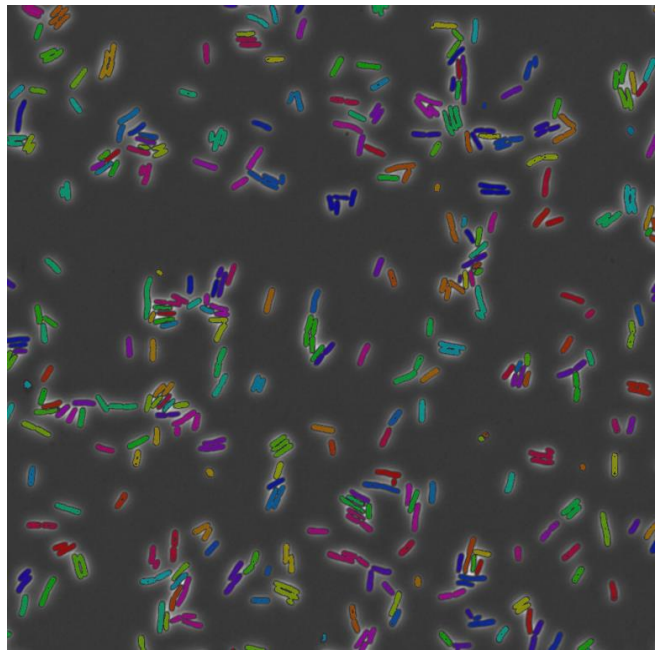

*Figure 4*: *An example for minimum thresholding finding all cells in the image in its segmentation*

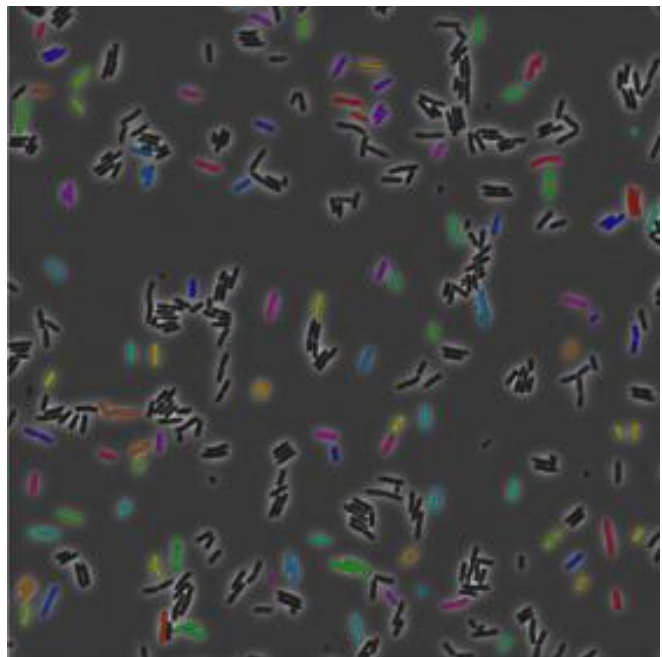

*Figure 5*: *An example of minimum thresholding segmentation with CellAnalyzer post-processing.*

4.1.2.1 **U-Net & Mask-RCNN**

Training for these models was conducted with 30 epochs with an initial learning rate of $1 \times 10^{-3}$ for the first 10 epochs, $1 \times 10^{-4}$ for the next 10 epochs, and $1 \times 10^{-5}$ for the last 10 epochs. The experimental $l_{ex}$-error is not a known method for segmentation evaluation, and thus is not offered as an option for training models in major deep-learning frameworks. In order to account for this logically using tried and tested evaluation metrics, the loss minimized for U-Net was binary cross-entropy loss summed with Jaccard loss to maximize overlap. These losses are defined as follows: let $pr \in M_{n \times m}([0,1])$ be a prediction probability map for an image whose ground truth is $G \in M_{n \times m}(\{0,1\})$. Given a matrix of size $A \in M_{n \times m}(S)$ (where $S$ is some value set), let $A^*$ be the flattening of matrix $A$, i.e. $A^* \in S^{n*m}$ and $A[i,j] = A^*[i*n + j]$. Then:

Binary Cross-entropy Loss:

$$H(pr, G) = -\frac{1}{n*m} \sum_{i=1}^{n*m} G^*[i] \cdot \log(pr^*[i]) + (1 - G^*[i]) \cdot \log(1 - pr^*[i])$$

Jaccard Loss:

$$d_J(pr, G) = 1 - \frac{\sum_{ij} pr[i,j] \cdot G[i,j]}{\sum_{ij} pr[i,j] + G[i,j] - \sum_{ij} pr[i,j] \cdot G[i,j]}$$

Mask-RCNN is a multi-task network which minimizes a specific set of losses, including loss for detection boxes which this experiment does not require. All networks losses were optimized using an Adam optimizer [33]. Images were fed to the models with a batch size of 1 (a single image every time) and every time an image is loaded it is randomly transformed using rigid transformations, e.g. flip (horizontal/vertical), rotate, transpose, etc. and brightness and gamma transformations.

Note: The exact model architectures used can be seen in appendix A.2.

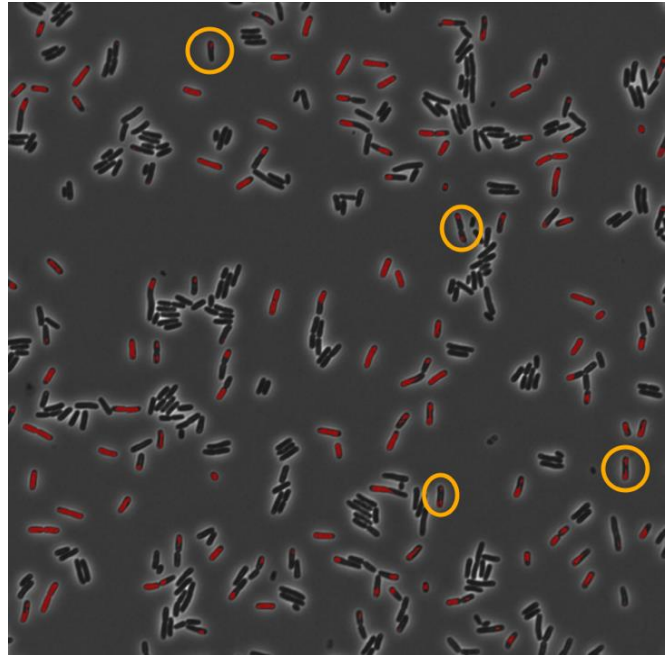

*Figure 6: An example of a U-Net segmentation with incomplete cell masks.*

A quick sanity check reveals that the U-Net model outputs a large number of incomplete segmentations (see *Figure 6*), while Mask-RCNN provides a clean output. This is justified by the fact that Mask-RCNN minimizes several losses aside from the mask, helping it concentrate on segmenting areas where cells have been detected. However, this U-Net issue is easily defeated by removing objects of a certain size from the segmentation, as done in CellAnalyzer.

Finally, looking at the experimental $l_{ex}$-errors of these models, it is clear that only Mask-RCNN meets the evaluation criteria for the purposes of this experiment (see *Table 1*).

4.1.2.2. **Cellpose.**

Cellpose attempts to generalize the cell segmentation task to many different kinds of cells and image formats. It can be seen clearly that even without extra training, the pretrained weights provided by the tool offer a visually seeming high-quality segmentation for this experiment's images.

| Cell Detection on Validation Image 1 | | | | | | | | | | | | | | | |
|---|---|---|---|---|---|---|---|---|---|---|---|---|---|---|---|
| $\beta$: 0.7 | $d_{ex}$: 0.174 | | | Researcher 1 Comparison | | | | | | Researcher 2 Comparison | | | | | |
| Model | TP | FP | FN | Precision | Recall | F2-Score | $l_{ex}$ | TP | FP | FN | Precision | Recall | F2-Score | $l_{ex}$ | Avg. $l_{ex}$ |
| Blank Mask | 0 | 0 | 155 | 0 | 0 | 0 | 0.275 | 0 | 0 | 122 | 0 | 0 | 0 | 0.216 | 0.246 |
| Thresholding (no CellAnalyzer) | 155 | 245 | 0 | 0.388 | 1 | 0.76 | 1.014 | 122 | 277 | 0 | 0.306 | 1 | 0.688 | 1.147 | 1.081 |
| Thresholding | 81 | 34 | 74 | 0.704 | 0.523 | 0.551 | 0.272 | 83 | 31 | 39 | 0.728 | 0.68 | 0.689 | 0.198 | 0.235 |
| U-Net | 33 | 3 | 112 | 0.934 | 0.277 | 0.323 | 0.211 | 46 | 3 | 76 | 0.938 | 0.377 | 0.428 | 0.147 | 0.179 |
| Mask-RCNN | **69** | **8** | **86** | **0.896** | **0.445** | **0.495** | **0.185** | **68** | **11** | **54** | **0.861** | **0.557** | **0.6** | **0.141** | **0.163** |
| Cell Detection on Validation Image 2 | | | | | | | | | | | | | | | |
| $\beta$: 0.7 | $d_{ex}$: 0.123 | | | Researcher 1 Comparison | | | | | | Researcher 2 Comparison | | | | | |
| Model | TP | FP | FN | Precision | Recall | F2-Score | $l_{ex}$ | TP | FP | FN | Precision | Recall | F2-Score | $l_{ex}$ | Avg. $l_{ex}$ |
| Blank Mask | 0 | 0 | 155 | 0 | 0 | 0 | 0.261 | 0 | 0 | 122 | 0 | 0 | 0 | 0.272 | 0.267 |
| Thresholding (no CellAnalyzer) | 155 | 245 | 0 | 0.388 | 1 | 0.76 | 0.345 | 122 | 277 | 0 | 0.306 | 1 | 0.688 | 0.309 | 0.327 |
| Thresholding | 81 | 34 | 74 | 0.704 | 0.523 | 0.551 | 0.328 | 83 | 31 | 39 | 0.728 | 0.68 | 0.689 | 0.271 | 0.3 |
| U-Net | 33 | 3 | 112 | 0.934 | 0.277 | 0.323 | 0.126 | 46 | 3 | 76 | 0.938 | 0.377 | 0.428 | 0.132 | 0.129 |
| Mask-RCNN | **69** | **8** | **86** | **0.896** | **0.445** | **0.495** | **0.126** | **68** | **11** | **54** | **0.861** | **0.557** | **0.6** | **0.120** | **0.123** |

*Table 1*: Cell detection validation results performed on validation images segmented by two different researchers. Only **Mask-RCNN** model upholds the experiment criterion giving valid predictions for all validation data. Note that the "blank mask" model should receive a better precision, recall and F2-Score than "thresholding (no CellAnalyzer)" model. The experiment assumptions and the researcher feedback say this is not the case. Only $l_{ex}$-error reflects this correctly.

However, this does not match the experiment assumptions since many cells that were not segmented by the researchers are segmented by Cellpose, including crowded cells and blurry ones. Furthermore, a "clump" of crowded cells that is useless for this experiment may be segmented as a single cell, which is the worst possible outcome under the assumptions, since it seems like one false positive detection, when actually it is several.

Training the model is not possible in the manner described for the previous models. The library containing this model was released with full documentation two months before the writing of this article. The user is given a choice of an initial learning rate and a number of epochs to run. For the rest of the training, the learning rate stays the same until the last 100 epochs where it is halved and is slowly deteriorated by a linear weight decay of $1 \times 10^{-5}$. The minimized loss is a sum of binary cross-entropy and L2 loss optimized by a standard SGD optimizer with momentum. Furthermore, the API does not allow access to the model during training, meaning that recording other metrics is not possible as of the writing of this paper.

Training with and without the provided weights seem to generalize well during training. However, without using the pretrained weights the model overfits on the last 100 epochs when the weight decay kicks in, and with the pretrained weights the model still segments "clumps" of cells. This is similar to the issues observed in using the Thresholding model.

4.2 **Fluorescence cluster segmentation.**

Delving into the fluorescence images data, one can see a recurring shape of a three-dimensional Gaussian distribution at its location, i.e. the algorithms should search for a three-dimensional Gaussian shape in the image (see *Figure 7*).

For this task, the evaluated models were U-Net and FPN. The choice to use only segmentation models instead of mask-RCNN detection model arises from mask-RCNN's poor performance on small objects.

Once again, the desired network is one that is produces valid segmentations relative to the validation set, but this time the experimental $l_{ex}$-error is defined mostly by the number of missing clusters.

| | | | | Fluorescence Cluster Segmentation on Validation Image 1 | | | | | | | | | | | | |
|---|---|---|---|---|---|---|---|---|---|---|---|---|---|---|---|---|
| $\beta$: | 0.15 | $d_{ex}$: | 0.412 | | Researcher 1 Comparison | | | | | | Researcher 2 Comparison | | | | | |
| | Model | | | TP | FP | FN | Precision | Recall | F2-Score | $l_{ex}$ | TP | FP | FN | Precision | Recall | F2-Score | $l_{ex}$ | Avg. $l_{ex}$ |
| | U-Net | | | 31 | 96 | 1 | 0.244 | 0.968 | 0.607 | 0.112 | 46 | 52 | 39 | 0.469 | 0.541 | 0.525 | 0.301 | 0.206 |
| | FPN | | | 31 | 198 | 1 | 0.135 | 0.968 | 0.434 | 0.224 | 68 | 51 | 37 | 0.571 | 0.647 | 0.631 | 0.287 | 0.256 |
| | | | | Fluorescence Cluster Segmentation on Validation Image 2 | | | | | | | | | | | | |
| $\beta$: | 0.15 | $d_{ex}$: | 0.174 | | Researcher 1 Comparison | | | | | | Researcher 2 Comparison | | | | | |
| | Model | | | TP | FP | FN | Precision | Recall | F2-Score | $l_{ex}$ | TP | FP | FN | Precision | Recall | F2-Score | $l_{ex}$ | Avg. $l_{ex}$ |
| | U-Net | | | 56 | 0 | 4 | 1 | 0.933 | 0.946 | 0.039 | 55 | 2 | 28 | 0.965 | 0.662 | 0.706 | 0.280 | 0.159 |
| | FPN | | | 58 | 14 | 2 | 0.805 | 0.966 | 0.929 | 0.044 | 58 | 3 | 25 | 0.950 | 0.698 | 0.737 | 0.252 | 0.148 |

*Table 2*: *Fluorescence cluster segmentation validation results. performed on validation images segmented by two different researchers. Both tested models are capable of giving valid predictions for this experiment and can both be considered new researchers.*

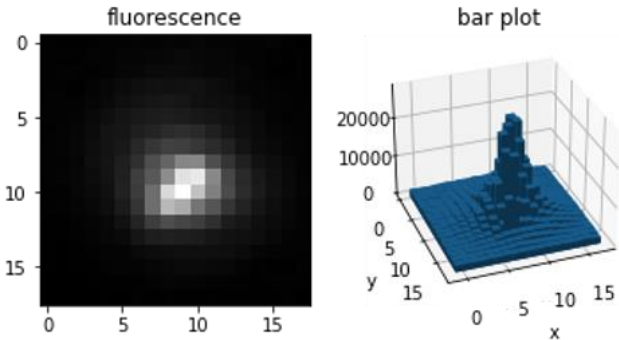

*Figure 7*: *An example of a single cluster image and 3D bar plot with matching axes (top left corners are (0,0)). Notice the general shape of a 3D Gaussian distribution plot.*

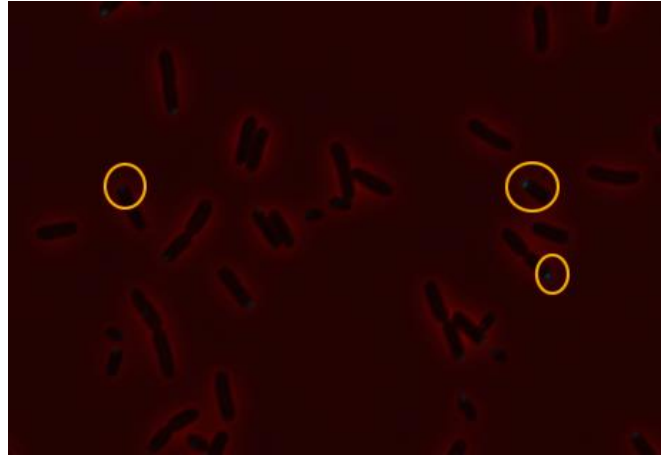

*Figure 8*: *A sample fluorescence cluster segmentation neural network input RGB image. The R channel contains the cells grayscale image and the G and B channels contain the fluorescence channel.*

U-Net and FPN were trained in the exact same way as the U-Net model for the cell segmentation task. In order to give the models a sense of the cell positions and to encourage finding clusters at the location of cells, the input for these models is an RGB image where the R channel is the cells image and the G and B channels are the fluorescence channel being segmented (see *Figure 8*).

U-Net and FPN both give valid predictions for both validation images (see *Table 2*). This means that these models are interchangeable for this experiment.

### 4.3 **Runtime**

#### 4.3.1 **Neural Network Training**

Model training was performed on a CUDA GPU. As mentioned earlier, all models were trained over 30 epochs. Mask-RCNN took approximately 2 minutes per epoch, and had a total runtime of 59.3 minutes. FPN took approximately 0.9 minutes per epoch, and had a total runtime of 27.8 minutes.

#### 4.3.2 **Analysis pipeline**

This flow was tested using two different hardware setups. The results can be seen in *Table 3*.

### V. DISCUSSION

#### 5.1. **Conclusion**

The objective of this study was to automate the process of cell and fluorescence channel analysis starting from the raw image output of the microscope and to bypass the cell segmentation bottleneck of today's tools. The tool was evaluated using the defined experimental $l_{ex}$-error and distance ($d_{ex}$) in section II. As seen in the Results section (IV), MicroAnalyzer truly does provide a good analysis for a valid number of cells and fluorescence clusters in each image, according to the defined evaluation method. This was done by using known object segmentation neural networks to find cells and fluorescence clusters in the image and Colicoords as an open-source alternative for analysis. This is a testament to the incredible flexibility and reliability of modern computer vision techniques, and how there might just be a model out there that can fit any experiment's data.

| Setup | Specifications | | | Runtime (seconds) | | | | | | |
|---|---|---|---|---|---|---|---|---|---|---|
| | OS | CPU | Memory | Read ND2 (per frame) | Mask-RCNN (per frame) | FPN (per frame) | Post-Processing (per cell) | Calculate Database Fields (per cell) | Generate Visualizations (per frame) | Full Pipeline (per frame) |
| Setup 1 | MacOS 10.14: High Sierra | Intel(R) Core™ i7-4870HQ CPU @ 2.50GHz | 16 GB of 1600 MHz DDR3 | 0.269 | 6 | 16 | 1.1 | 0.013 | 1.5 | 100 |
| Setup 2 | Windows 10 (OS build 19041.450) | Intel(R) Core™ i7-8565U CPU @ 1.80GHz | 16GB of 2400 MHz DDR4 | 0.34 | 3.935 | 28.1 | 1.07 | 0.009 | 1.503 | 120 |

*Table 3: Runtime results table for two different hardware specifications and operating systems.*

Nevertheless, the segmentation evaluation method used is very specific to the experiment discussed in this paper. The models used here may not be as efficient for other experiment assumptions and images. Cellpose and others like it may be key in any generic version of MicroAnalyzer.

5.2. **Future Work**

One option for the expansion of this project is analyzing three-dimensional cell images. Cellpose offers a feature that finds cells in 3D microscopic images, and thus may be a viable option for this task. The experiment has a certain emphasis on polar localization, but MicroAnalyzer doesn't actually require any other channels but the cell images. Without fluorescence data, the output database and visualizations still contain cell segmentation and analysis data, which may be useful for experiments that do not rely on additional channels. Something similar could be achieved with three-dimensional images as well.

Time-lapse image analysis is another form of microscopic output used for studying the organisms' behavior and subcellular organization over time. Specifically, for this experiment, the time-lapse images' time data exists within the ND2 files. Object tracking networks are available in open-source repositories and are proving reliable, making tasks such as tracking cell mitosis frequency seem undaunting as a logical next step for the development of MicroAnalyzer.

Another possible direction is the analysis of different localization patterns other than polar. Patterns like the helix can be far more difficult to detect as they do not have the signature Gaussian shape and should not be properly segmented using MicroAnalyzer's models. Perhaps the correct way to do this is to look at each cell individually and to classify the localization of the protein in the cell. The experiment in this paper concentrates on polar localization, but there very may well be a demand for other localizations in the future.

5.3. **Further Discussion**

The ability to perform cell and fluorescence cluster analysis quickly and with minimal human interaction will allow labs to produce enormous amounts of analysis data in a much shorter time, and even shorter as the lab upgrades their hardware. Statistical questions that could not be answered previously due to lack of data can now be studied more deeply, e.g. perhaps a certain set of properties of a cell and its fluorescence data point to some phenomenon with a high probability. Neural networks for regression and classification are often used for these tasks, and in this case create a chain of neural networks working together to achieve one larger goal. Now imagine automating the entire pipeline: the microscope takes thousands of images overnight which are input into MicroAnalyzer to generate data for hundreds of thousands of cells and the studied material, and run the statistical analysis algorithm on this giant database. Even if this takes a month to run, it is much faster than performing a manual experiment filled with pesky, unpredictable human errors. It also frees the researchers to perform other tasks that cannot (yet) be performed reliably by a machine. Herein lies the true power of the dynamic duo that is machine learning and automation.

## Acknowledgements

The idea for this project came from members of Orna Amster-Choder lab, Tamar Szoke, Nitsan Albocher and Omer Goldberger, who raised the need for a tool to analyze their fluorescence microscopy data. We thank them for putting the time to define their needs, provide fluorescence images and analyze them manually.

# APPENDICES

**Appendix A: Relevant Links**

1. MicroAnalyzer Repository
   - https://github.com/JG-codies/MicroAnalyzer
2. Neural Network Model Architectures
   - https://github.com/JG-codies/MicroAnalyzer/blob/master/Notebooks/model_summaries.ipynb
3. Dataset
   - https://drive.google.com/drive/folders/1byIX3DtaSTsBLF8a91012ljtrNkqVGa8?usp=sharing
4. Neural Network Models Training[8]
   - Cells – https://github.com/JG-codies/MicroAnalyzer/blob/master/Notebooks/cell_training.ipynb
   - Clusters – https://github.com/JG-codies/MicroAnalyzer/blob/master/Notebooks/fluo_training.ipynb
5. Validation
   - Image 1 – https://github.com/JG-codies/MicroAnalyzer/blob/master/Notebooks/Validation%20Image%201.ipynb
   - Image 2 – https://github.com/JG-codies/MicroAnalyzer/blob/master/Notebooks/Validation%20Image%202.ipynb
6. Usage Demo
   - https://github.com/JG-codies/MicroAnalyzer/blob/master/Notebooks/usage_demo.ipynb
7. Sample DB
   - https://github.com/JG-codies/MicroAnalyzer/blob/master/sample_output/validation_images/1/database.csv

**Appendix B: "Soft" Segmentation Evaluation**

Remember that since the experiment is evaluated using multiple ground truth segmentations, the definition the evaluation method in section III considers a *valid* prediction as if it were a segmentation from another researcher. However, what if there are extra cells that appear in the prediction that do not appear in either ground truth segmentation? Should it be considered as a "worse" false positive? Could it possibly be a true positive that a third researcher might have segmented?

---
[8] Performed in "Google Colab" with GPU. Training details are described int the article (Results section).

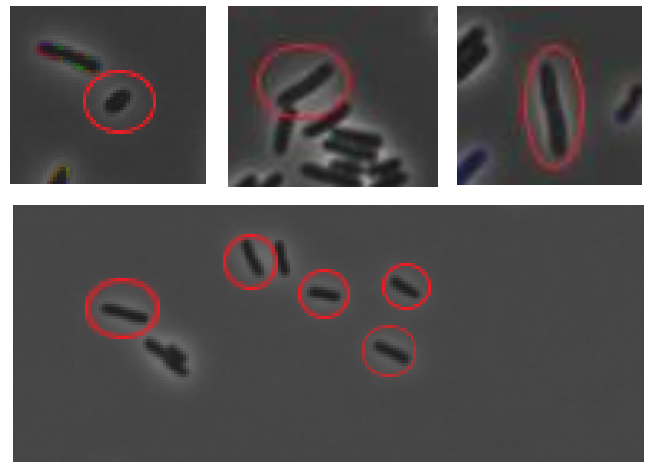

*Figure 9*: Cells detected by Maks-RCNN that appeared in only one of the two ground truth segmentations due to disagreement on size, spacing, and focus.

The reason such a prediction should not be additionally penalized for this is because it is already penalized twice for each image in the evaluation criterion (once for each researcher segmentation), and more importantly it may be a valid cell that both researchers missed due to human faults. Once the rule of multiple ground truths has been accepted, the experiment relies on human competence which can never be guaranteed, and thus all result outcomes are probabilistic and not deterministic, forcing this last evaluation to be performed manually by the researchers. This helps give a general idea of how well the model fits to the desires of the researchers. In the manual check of the predictions of the Mask-RCNN model on the validation set, the researchers debated amongst themselves about the validity of the "rogue" cell detections (as seen in *Figure 9*), showing that keeping those cells for analysis may or may not be valid. This phenomenon may hint as to why deep-learning models greatly outperformed the deterministic image processing approach.

It is irrelevant to talk about cells that appear in both ground truths but are missing in the prediction since the already defined evaluation guarantees that not "too many" cells are missing (depending on the chosen $\beta$).

**Appendix C: Preprocessing**

The input file is an ND2 file which is a Nikon proprietary file type containing a set of 16bit microscope images with all the camera's channels, including the grayscale bacteria image and the fluorescence intensity channels.

Given an ND2 file, initially separate the bacteria image channel from the fluorescence channels. The bacteria image pre-processing is as follows:

1. Convert the image to 8bit RGB image (all channels are the same).

2. Pad the images such that they're dimensions are divisible by $2^5$. This allows us to down-sample the image evenly at least five times (required by the neural networks).

Each fluorescence image is pre-processed in the same way as the bacteria image, except that the R channel of the image is substituted with the 8bit version of the bacteria image.

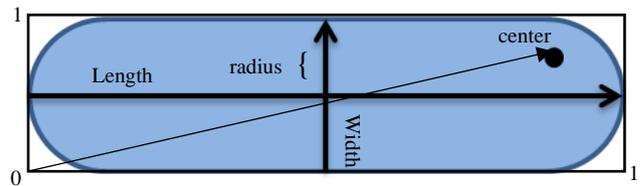

*Figure 10*: A partial database columns description diagram,

**Appendix F: API**

MicroAnalyzer was built to support ND2 files and the name of the channel containing the cell images as inputs and outputs a database with the results of the analysis and several visualizations of the segmentations.

A. **Id** – the identifier in the mask of the ROI presented in the row.
B. **frame id** – the index of the image in the nd2 file that the ROI was found in.
C. **length** – the ROI length in $\mu m$ (see *Figure 10*).
D. **width** – the ROI width in $\mu m$ (see *Figure 10*).
E. **area** – the 2D area of the ROI according to the image in $\mu m^2$.
F. **radius** – the distance between the edge of the ROI and its mid-line in $\mu m$ (see *Figure 10*).
G. **circumference** – the length of the perimeter of the ROI in $\mu m$.
H. **surface area** – an estimation of the surface area of the ROI modeled as a 3D object.
I. **Volume** – an estimation of the volume of the ROI modeled as a 3D object.

J. **<Fluorescence-name> cell mean/std intensity** – the mean/std pixel intensity of the fluorescence in the entire boundary of the cell.
K. **<Fluorescence-name> cell intensity CVI** – "<Fluorescence-name> cell mean intensity" divided by <Fluorescence-name> cell std intensity (both calculated by Colicoords).
L. **<Fluorescence-name> vertical/horizontal mean/max/sum intensity profile** – sample 20 points evenly along the vertical/horizontal axis and aggregate the fluorescence intensity on the perpendicular axis according to the function mean/max/sum.

M. **<Fluorescence-name> number of clusters** – the number of clusters that intersect the cell boundary (pixel-wise).
N. **<Fluorescence-name> has clusters** - a Boolean that is true if and only if the matching "number of clusters" field is not 0.
O. **<Fluorescence-name> cluster <index> id** - the identifier in the mask of cluster <index> presented in the row.
P. **<Fluorescence-name> cluster <index> size** – the size of the cluster in $\mu m^2$ according to cluster mask in image.
Q. **<Fluorescence-name> cluster <index> center** – a tuple (x, y) of numbers between 0 and 1 representing the position of the cluster (the point of max intensity) in proportion to the boundaries of the cell (see *Figure 10*).
R. **<Fluorescence-name> cluster <index> is polar** – A boolean that is true if and only if the cluster center x coordinate is less than 0.25 or greater than 0.75.
S. **<Fluorescence-name> cluster <index> mean/std/max/sum intensity** – the mean/std/max/sum of the fluorescence image pixel intensity within the boundaries of the cluster.
T. **<Fluorescence-name> leading cluster index** – the index of the cluster (that appears in the column headers) with the highest "max intensity" field.